%% file: main.tex
\documentclass[final,12pt]{colt2024} 

\title[Kernel-Based RL]{Open Problem: Order Optimal Regret Bounds for Kernel-Based Reinforcement Learning}
\usepackage{times}

\coltauthor{%
 \Name{Sattar Vakili} \Email{sv388@cornell.edu}\\
 \addr MediaTek Research
}

\input{notation}

\begin{document}

\maketitle

\begin{abstract}%
  Reinforcement Learning (RL) has shown great empirical success in various application domains. The theoretical aspects of the problem have been extensively studied over past decades, particularly under tabular and linear Markov Decision Process structures. Recently, non-linear function approximation using kernel-based prediction has gained traction. This approach is particularly interesting as it naturally extends the linear structure, and helps explain the behavior of neural-network-based models at their infinite width limit.  The analytical results however do not adequately address the performance guarantees for this case. We will highlight this open problem, overview existing partial results, and discuss related challenges.   
\end{abstract}

\begin{keywords}%
  Kernel methods, reinforcement learning, order optimal performance guarantees  
\end{keywords}

\input{text}


\bibliography{ref}

\appendix






\end{document}

%% file: notation.tex
\usepackage{bm}
\usepackage{amsmath}
\usepackage{amssymb}
\usepackage{mathtools}

\usepackage{booktabs} 
\usepackage{hyperref} 

\usepackage{tikz}
\usetikzlibrary{automata, positioning, arrows.meta}

\tikzset{
    state/.style={
        circle,
        draw,
        minimum size=1cm
    }
}

\def\Hc{\mathcal{H}}
\def\Sc{\mathcal{S}}
\def\Ac{\mathcal{A}}
\def\Zc{\mathcal{Z}}

\def\Rc{\mathcal{R}}
\def\Oc{\mathcal{O}}

\def\Vc{\mathcal{V}}
\def\Nc{\mathcal{N}}

\def\Oc{\mathcal{O}}
\def\Oct{\tilde{\mathcal{O}}}

\def\E{\mathbb{E}}

\def\Rr{\mathbb{R}}

\newtheorem{assumption}{Assumption}

%% file: text.tex
\section{Introduction}

The analytical study of Reinforcement Learning (RL) faces a natural progression in the complexity of the Markov Decision Process (MDP) structure: Tabular $\rightarrow$ Linear $\rightarrow$ Kernel-based $\rightarrow$ Neural-network-based. Kernel-based models serve as natural intermediary between well-studied linear models and the less understood neural-network-based models~\citep[having the neural tangent kernel theory in mind; see, e.g., the discussions in][]{yang2020provably}. Kernel-based models provide a rich representation capacity for nonlinear function approximation in RL, while still lending themselves to theoretical analysis, and have gained traction in recent years. However, some fundamental problems still remain open.

For a sharp and clear presentation of our open problem, we focus on an episodic MDP within an online framework under the regret performance measure. However, similar problems can be raised in other settings, including infinite horizon discounted or undiscounted MDPs within an online or offline framework. We specifically ask:
\emph{Is it possible to design order-optimal or, at the very least, no-regret learning algorithms under reasonable assumptions on an MDP with a kernel-based structure?}
In this paper, we will formally present this open problem, provide an overview of existing work, and discuss some of the challenges involved.



\section{Episodic MDP}\label{sec:MDP}

An episodic MDP can be described by the tuple $M=(\Sc,\Ac, H, P, r)$, where $\Sc$ and $\Ac$ are the state and action spaces, the integer $H$ is the length of each episode, $r=\{r_h\}_{h=1}^H$ are the reward functions and $P=\{P_h\}_{h=1}^H$ are the transition probability distributions.
We use the notation $\Zc=\Sc\times\Ac$ to denote the state-action space. For each $h\in[H]$, $r_h: \Zc\rightarrow [0,1]$ is the reward function at step $h$, which is supposed to be deterministic and known for simplicity, and $P_h(\cdot|s,a)$ is the \emph{unknown} transition probability distribution on $\Sc$ for the next state from state-action pair~$(s,a)$.



A policy $\pi=\{\pi_h\}_{h=1}^H$, at each step $h$, determines the (possibly random) action $\pi_h:\Sc\rightarrow \Ac$ taken by the agent at state $s$.  
At the beginning of each episode $t=1,2,\cdots$, the environment picks an arbitrary state $s_{1,t}$. The agent determines a policy $\pi_{t}=\{\pi_{h,t}\}_{h=1}^H$. Then, at each step $h\in[H]$, the agent observes the state $s_{h,t}\in\Sc$, and picks an action $a_{h,t}=\pi_{h,t}(s_{h,t})$. The new state $s_{h+1,t}$ then is drawn from the transition distribution $P_h(\cdot|s_{h,t}, a_{h,t})$. The episode ends when the agent receives the final reward $r_H(s_{H,t},a_{H,t})$.  
We are interested in maximizing the expected total reward in the episode, starting at step $h$, i.e., the value function, defined as:
\begin{equation}
V^{\pi}_h(s) = \E\left[\sum_{h'=h}^Hr_{h'}(s_{h'},a_{h'})\bigg|s_{h}=s\right],  ~~~\forall s\in\Sc, h\in[H],
\end{equation}
where the expectation is taken with respect to the randomness in the trajectory $\{(s_h,a_h)\}_{h=1}^H$ obtained by the policy~$\pi$. It can be shown that under mild assumptions (e.g., continuity of $P_h$, compactness of $\Zc$, and boundedness of $r$) there exists an optimal policy $\pi^{\star}$ which attains the maximum possible value of $V^{\pi}_{h}(s)$ at every step and at every state~\citep[e.g., see,][]{puterman2014markov}. We use the notation
$
V_{h}^{\star}(s) = \max_{\pi}V_h^{\pi}(s), ~\forall s\in\Sc, h\in[H]
$.
By definition $V^{\pi^{\star}}_h=V_{h}^{\star}$.
The performance of a learning algorithm $\{\pi_t\}_{t\in[T]}$ is measured in terms of the total loss in the value function, referred to as \emph{regret}, denoted by $\Rc(T)$ in the following definition:
\begin{equation}
\Rc(T) = \sum_{t=1}^T(V^{\star}_1(s_{1,t}) - V_1^{\pi_t}(s_{1,t})).
\end{equation}
A learning algorithm with sublinear regret in $T$ is often referred to as a \emph{no-regret} algorithm, since the average regret over $T$ tends toward zero as $T$ increases. This implies that the value of the policy executed by the learning algorithm converges to that of the optimal policy over episodes.

For a value function $V:\Sc\rightarrow\Rr$, and a conditional distribution $P(s|z)$, $s\in\Sc, z\in\Zc$, we define the notation $[PV](z)=\E_{s\sim P(\cdot|z)}[V(s)]$.
The state-action value function $Q^{\pi}_h:\Zc\rightarrow [0,H]$ is defined as follows:
$
Q_h^{\pi}(s,a) = \E_{\pi}\left[
\sum_{h'=h}^Hr_{h'}(s_{h'},a_{h'})\bigg|s_h=s, a_h=a
\right],
$
where the expectation is taken with respect to the randomness in the trajectory $\{(s_h,a_h)\}_{h=1}^H$ obtained by the policy~$\pi$.
The Bellman equation associated with a policy $\pi$ then is represented as:
$
Q_h^{\pi}(s,a) = r_h(s,a) + [P_hV^{\pi}_{h+1}](s,a)$,
$V_h^{\pi}(s) = \max_{a\in\Ac}Q_h^{\pi}(s,a)$, $
V_{H+1}^{\pi} \equiv \bm{0}$.


\section{Kernel-Based Modelling}\label{sec:kernel}

Various structural complexities for MDPs have been considered, including the tabular model with small finite $\mathcal{S}$ and $\mathcal{A}$, where regret bounds of $\Oct(\sqrt{|\Sc||\Ac|H^3 T})$ have been shown~\citep[see, e.g.,][]{jin2018q}. In the linear setting, the transition probability model $P_h$ is assumed to be representable using a linear feature mapping~\citep{jin2020provably}: 
\begin{equation*}
    P_h(\cdot|s,a) = \bm{\theta}_h(\cdot)\bm{\phi}(s,a), ~~~ \forall s \in \Sc, a\in\Ac,
\end{equation*}
where $\bm{\theta}_h(\cdot)\in \Rr^d$ are unknown measures over $\Sc$, and $\bm{\phi}:\Sc\times\Ac\rightarrow \Rr$ is a $d$-dimensional feature map.
This representation enables the use of linear function approximation for the expected value function $[P_hV]$ and facilitates the design of a policy based on an optimistic modification of Least-Squares Value Iteration (LSVI) that achieves a regret bound of $\Oct(\sqrt{d^3H^3T})$, scaling with the dimension $d$ of the feature map $\bm{\phi}$, rather than the size of $\Sc$ and $\Ac$.

A natural extension of the linear model is the kernel-based model, which is a linear model in the (possibly infinite-dimensional) feature space of a positive definite kernel. This approach provides rich representation capacity for nonlinear function approximation using kernel-ridge regression. 
An intuitive explanation of this approach can be provided by using Mercer's theorem. Let $k:\Zc\times\Zc\rightarrow \Rr$ be a known positive definite kernel. By Mercer's theorem, $k$ can be expressed as $
k(z,z') = \sum_{m=1}^\infty\lambda_m\varphi_m(z)\varphi_m(z')$,
where $\lambda_m > 0$ and $\varphi_m : \Zc \rightarrow \Rr$ are the eigenvalues and eigenfeatures corresponding to $k$, respectively. Additionally, the space defined by these functions is referred to as the reproducing kernel Hilbert space (RKHS) corresponding to $k$:
\begin{equation*}
    \Hc_k = \left\{ f: f=\sum_{m=1}^\infty \theta_m \lambda_m^{\frac{1}{2}}\varphi_m(z), ~ \|\bm{\theta}\|<\infty \right\},
\end{equation*}
where $\|\bm{\theta}\|$ is the $\ell_2$ norm of the weights $\theta_m$. The RKHS norm of $f$ is defined as $\|f\|_{\Hc_k}=\|\bm{\theta}\|$.
I.e., $\Hc_k$ is the class of linear functions in the feature space of $\{\lambda_m^{\frac{1}{2}} \varphi_m\}_{m=1}^{\infty}$.

Analogous to the linear model, the following assumption is made for kernel-based models, with the linear model as a special case when using a linear kernel.
\begin{assumption}\label{ass:kernel}
    Let $\Hc_k$ be the RKHS corresponding to a positive definite kernel $k:\Zc\times\Zc\rightarrow \Rr$. We assume, $\forall h\in[H], \forall s'\in\Sc$, $P_h(s'|\cdot)\in\Hc_k$ and $\|P_h(s'|\cdot)\|_{\Hc_k}\le u$, for some constant $u>0$. 
\end{assumption}
The target function of interest here is $f=[PV]$ for some unknown transition probability distribution $P$ satisfying Assumption~\ref{ass:kernel}, and some value function  $V:\Sc\rightarrow \Rr$. For a dataset of $n$ transitions $\{(z_i,s'_i)\}_{i=1}^n$, where $s'_i\sim P(\cdot|z_i)$, we have the following prediction and uncertainty estimate, respectively, for $f$, utilizing kernel-ridge regression:
\begin{equation}\label{eq:krr}
    \hat{f}_n(z) = \bm{k}^{\top}_n(z) (\bm{K}_n+\rho I)^{-1}\bm{v}_n,
    ~~~ \text{and} ~~ \sigma_n^2(z) = \bm{k}^{\top}_n(z) (\bm{K}_n+\rho I)^{-1}\bm{k}_n(z),
\end{equation}
where $\bm{k}_n=[k(z,z_1), k(z,z_2), \cdots, k(z,z_n)]$, $\bm{K}_n=[k(z_i,z_j)]_{i,j=1}^n$ is the kernel matrix, $\rho>0$ is a free parameter, $I$ is the identity matrix of dimensions $n$, and $\bm{v}_n=[V(s'_1), V(s'_2), \cdots, V(s'_n)]$ is the vector of observations. Note that $\E_{s'\sim P(\cdot|z)}[V(s')]=f(z)$ by definition; thus, $V(s'_i)$ are noisy realizations of $f(z_i)$.

\section{Open Problem}
The question we ask is as follows:
\emph{Consider the episodic MDP setting described in Section~\ref{sec:MDP}. Under Assumption~\ref{ass:kernel}, (a) Can a no-regret learning algorithm be designed? 
(b) What is the minimum regret growth rate with $T$ (and also $H$)? And, can a learning algorithm be designed to achieve order-optimal (or near-optimal) regret performance, closely aligning with the established lower bound?}

\section{Existing Results and Challenges}

Recall the kernel ridge prediction $\hat{f}_n$ and the uncertainty estimate $\sigma_n$ given in Equation~\eqref{eq:krr}. Confidence intervals based on these statistics are important building blocks in the analysis of function approximation in RL using kernel-based models. 
For a fixed $f\in \Hc_k$, with $\|f\|_{\Hc_k}\le c_f$, for some constant $c_f>0$, assuming $\sigma$-sub-Gaussian observation noise, several confidence intervals of the form $|f(z)-\hat{f}_n(z)|\le \beta_n(\delta)\sigma_n(s)$, at a $1-\delta$ confidence level
are established, where the confidence interval width multiplier $\beta_n(\delta)$ depends on $\delta$, $n$, $\sigma$, and the kernel $k$~\citep{abbasi2013online,
chowdhury2017kernelized, vakili2021optimal,  whitehouse2024sublinear}. 
In particular, in an offline setting where the observation points $z_i$ are predetermined and independent of the observation noise, the confidence interval for a fixed $z\in\Zc$ is given by $\beta_n(\delta) = c_f + \frac{\sigma}{\sqrt{\rho}}\sqrt{2\log(\frac{1}{\delta})}$~\citep{vakili2021optimal}. In an online setting, such as RL and bandits, where the observation points are adaptively selected based on prior observations, the application of self-normalized confidence bounds for vector-valued martingales~\citep{abbasi2013online} to the kernel setting results in a confidence interval with $\beta_n(\delta) = c_f+\frac{\sigma}{\sqrt{\rho}}\sqrt{2\log(\frac{1}{\delta})+\gamma(n)}$~\citep{whitehouse2024sublinear}, where $\gamma(n)=\sup_{z_1, \cdots, z_n\in\Zc}\frac{1}{2}\log\det(I+\rho^{-1}\bm{K}_n)$ represents the maximum information gain, a kernel specific complexity term that intuitively reflects the \emph{effective dimension} of the kernel model. 

In the RL setting, function approximation is typically applied to $f_{h,n}=[P_h V_{h,n}]$, with proxies $V_{h,n}:\Sc \rightarrow \Rr$ for the value function---usually an upper confidence bound on the value function. Due to the Markovian
nature of the temporal dynamics, these functions are not prefixed, making the previously mentioned confidence intervals inapplicable. This necessitates further considerations, leading to an increased $\beta_n(\delta)$. We here overview the most important approaches addressing this issue and the corresponding eventual regret bounds. 

In~\cite{yang2020provably}, the authors adopt a rigorous approach, considering the class of all proxy value functions that appear throughout their optimistic LSVI algorithm:
\begin{equation*}
    \Vc = \{V: \forall (s,a)\in\Zc, V(s)=\max_{a\in\Ac}\min\{H, Q(s,a)\}, Q(z) = Q_0(z) + b\sigma_n(z) \},
\end{equation*}
where $\|Q_0\|_{\Hc_k}\le c$, for some $c>0$, and $b\in[0,B]$, $Q_0$ represents possible predictions and $\sigma_n$ represents possible uncertainty estimates derived from a set of $n$ observations using kernel ridge regression. They proceed by bounding the $\epsilon$-covering number of $\Vc$. Specifically, they establish a bound on $\Nc(\Vc, \epsilon)$, the minimum number of functions needed to cover $\Vc$ up to an $\epsilon$ error in $\ell_\infty$ norm. Using this technique, they derive a confidence interval that is applicable to all $V\in\Vc$ and the corresponding $f=[PV]$, with $\beta_n(\delta)= \Oc(c_f+\sqrt{\log(\frac{1}{\delta}) + \gamma(n) + \log\Nc(\Vc, \frac{1}{n})})$ for a choice of $\epsilon=\Oc( \frac{1}{n})$. This leads to a regret bound of $\Oct(\beta_T(\delta)H^2\sqrt{T\gamma(T)})$, which does not ensure  no-regret performance when $\beta_T(\delta)$ grows at least as fast as $\sqrt{T}$, a case common with several kernels of interest, including the Met{\'e}rn and neural tangent kernels.

\cite{chowdhury2023value} make a strong assumption, referred to as \emph{optimistic closure}, that $\forall V\in\Vc, \|V\|_{\Hc_k}\le c_v$, for some $c_v>0$. Utilizing kernel mean embedding, they prove a tighter confidence interval with $\beta_n(\delta) = \Oct(c_f+\frac{\sigma}{\sqrt{\rho}} \sqrt{\log(\frac{1}{\delta})+\gamma(n)})$. This improved bound potentially leads to no-regret guarantees using techniques from bandits~\citep[see,][]{whitehouse2024sublinear}. However, the optimistic closure assumption seems unrealistic. One potential solution to rigorously obtain no-regret guarantees involves relaxing the optimistic closure assumption in their work.  

A related problem is kernel bandits (also known as Bayesian optimization), a spacial case with $|\Sc|=1, H=1$, where no-regret guarantees are established for Upper Confidence Bound type algorithms~\citep{whitehouse2024sublinear}, and more sophisticated algorithms that perform sample or domain partitioning have been shown to achieve order-optimal regret bounds~\citep[see, e.g.,][]{Valko2013kernelbandit, salgia2021domain, li2021gaussian}. While the ideas in these works are insightful, their applicability to general MDPs is unclear.

\newpage

